\documentclass[conference]{IEEEtran}
\IEEEoverridecommandlockouts
\usepackage{cite}
\usepackage{amsmath,amssymb,amsfonts}
\usepackage{algorithmic}
\usepackage{graphicx}
\usepackage[labelformat=empty]{subcaption}
\graphicspath{ {./images/} }
\usepackage{textcomp}
\usepackage{xcolor}
\def\BibTeX{{\rm B\kern-.05em{\sc i\kern-.025em b}\kern-.08em
    T\kern-.1667em\lower.7ex\hbox{E}\kern-.125emX}}
\begin{document}

\title{Exploring Regions of Interest: Visualizing Histological Image Classification for Breast Cancer using Deep Learning\\
\thanks{Identify applicable funding agency here. If none, delete this.}
}

\author{\IEEEauthorblockN{ Imane Nedjar }
\IEEEauthorblockA{\textit{
Biomedical Engineering Laboratory} \\
\textit{Higher School of Applied Sciences of Tlemcen }\\
Tlemcen, Algeria \\
imane.nedjar@univ-tlemcen.dz}
\and
\IEEEauthorblockN{Mohammed Brahimi}
\IEEEauthorblockA{\textit{Computer Science Department} \\
\textit{USTHB University }\\
Algiers, Algeria \\
mohamed.brahimi@univ-bba.dz}
\and
\IEEEauthorblockN{ Saïd Mahmoudi }
\IEEEauthorblockA{\textit{Computer Science Department} \\
\textit{University of Mons}\\
Mons, Belgium \\
said.mahmoudi@umons.ac.be}
\and
\IEEEauthorblockN{ Khadidja Abi Ayad }
\IEEEauthorblockA{\textit{ Department of Pathology} \\
\textit{University Hospital Center of Tlemcen }\\
Tlemcen, Algeria \\
Khadidja.AbiAyad@univ-tlemcen.dz}
\and
\IEEEauthorblockN{Mohammed Amine Chikh}
\IEEEauthorblockA{\textit{
Biomedical Enginnering Laboratory} \\
\textit{Tlemcen University}\\
Tlemcen, Algeria \\
ma-chikh@univ-tlemcen.dz}
\and
}

\maketitle

\begin{abstract}
Computer-aided detection and diagnosis systems (CADe/CADx) based on deep learning have shown promising performance in breast cancer detection. However, there are cases where the obtained results lack justification. In this study, our objective is to highlight the regions of interest used by a convolutional neural network (CNN) for classifying histological images as benign or malignant. We compare these regions with the regions identified by pathologists.
\\To achieve this, we employed the VGG19 architecture and tested three visualization methods: Gradient, LRP-Z, and LRP-Epsilon. Additionally, we experimented with three pixel selection methods: Bins, K-means, and MeanShift. Based on the results obtained, the Gradient visualization method and the MeanShift selection method yielded satisfactory outcomes for visualizing the images.

\end{abstract}


\section{Introduction}
According to the American Cancer Society in the United States, breast cancer deaths have continued to decrease in older women. This can be attributed to earlier diagnostics and improved treatments \cite{Ref1}. In fact, advancements have been made in the materials used for treatment, such as digital mammography and optical microscopy. Additionally, there has been significant development in Computer-Aided Diagnosis (CADx) systems. Notably, the CADx systems based on Convolutional Neural Networks (CNNs) have gained popularity due to their high precision.
\\However, in certain cases, medical experts find it challenging to understand the reasoning and decisions made by these systems. To address this issue, this study proposes the application of visualization methods to the classification of hematoxylin and eosin stained breast biopsy images using CNNs. Specifically, well-known visualization methods such as Gradient and Layer-Wise Relevance Propagation (LRP) are tested.
\\The paper is organized into five sections. Section 2 provides a brief review of visualization methods and their applications in the medical field. In section 3, the visualization and selection methods used in this study are introduced. Section 4 discusses the results of the experiments conducted. Finally, the paper concludes with a summary of findings and conclusions.
\section{Related works}
CNNs have been widely employed in various applications of medical image processing, including classification \cite{Ref2} \cite{Ref3} and segmentation \cite{Ref4} \cite{Ref5}. However, understanding the behavior of CNNs can be challenging, which has significant implications, particularly in sensitive domains like the medical field. In particular, when there is a discrepancy between the decisions made by CNNs and medical experts, the lack of transparency becomes crucial. Thankfully, in recent years, several interpreted and visualization methods for CNNs have been developed to address this issue and enhance transparency.
\\Simonyan et al. \cite{Ref6} employed the back-propagation pass through a CNN to compute the spatial support of a specific class within an image. This method allowed them to understand the regions in the image that contributed to the classification decision made by the CNN. 
\\On the other hand, Zeiler et al. \cite{Ref7} proposed a visualization method based on the deconvolutional network. This method involves reconstructing the input of each layer from its output using top-down projections. By doing so, the method reveals the structures within each patch of the image that stimulate a particular feature map. Additionally, the authors demonstrated the sensitivity of the CNN to local structures in the image by utilizing the occlusion method.
\\The layer-wise relevance propagation method, proposed by Bach et al. \cite{Ref8}, utilizes a propagation rule to distribute the class relevance found at a given layer back onto the previous layer. These visualization methods, including layer-wise relevance propagation, have been applied in the medical context to gain insights into the decisions made by CNNs. By employing these methods, medical professionals can better understand and interpret the reasoning behind CNN-based decisions.
\\For example, in the diagnosis of Alzheimer's disease, there have been studies that utilize visualization techniques to identify and select the regions in medical images that are associated with the disease \cite{Ref9}  \cite{Ref10}. By employing visualization methods, medical expert can highlight and analyze specific areas of interest in the brain that are indicative of Alzheimer's disease, aiding in the diagnostic process.
\\In \cite{Ref11}, the authors conducted a comparison between two families of visualization methods: (1) the gradient-based methods, including Sensitivity Analysis and Guided Backpropagation, and (2) the occlusion-based methods, such as Occlusion and Brain Area Occlusion. The study aimed to evaluate the effectiveness and performance of these visualization techniques. The authors of \cite{Ref11} concluded that both visualization methods, namely the gradient-based methods (Sensitivity Analysis, Guided Backpropagation) and the occlusion-based methods (Occlusion, Brain Area Occlusion), focus on the same region of the brain. However, they noted that there are certain differences between these methods in terms of the specific details and nuances revealed in the visualizations. Despite these differences, both techniques provide valuable insights into the region of interest in the brain for the given application.
\\In the field of dermatology, Molle et al. \cite{Ref12} employed feature maps to examine the features utilized by CNNs for skin lesion classification. The authors highlighted the similarity in concepts used by both CNNs and dermatologists, such as the lesion border.

In our work, our focus lies in the application of visualization methods to classify histological images. In the following section we will present the CNN and the visualization methods used.
\section{Materials and methods}
\subsection{Dataset}
Several publicly available datasets containing histopathological breast cancer images exist, including Bio-segmentation \cite{Ref13}, MITOS-ATYPIA-14 \cite{Ref14}, Janowczyk \cite{Ref15}, and BreakHis \cite{Ref16}. Considering the large number of images available in the BreakHis dataset, we have selected it for our evaluation. 
\\The BreaKHis database consists of microscopic biopsy images of both benign and malignant breast tumors. These samples are derived from breast tissue biopsy slides that have been stained with hematoxylin and eosin (HE). The database contains a total of 7,909 images, which are categorized into benign and malignant tumors. For evaluation purposes, the dataset provides five different training-testing splits, where each split allocates 70\% of the images for training and 30\% for testing \cite{Ref16}. 
\begin{figure}[!ht]
 \centering
 \includegraphics[width=0.5\textwidth]{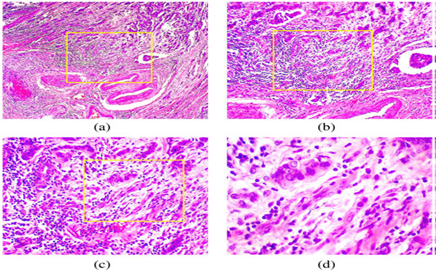}
  \caption{Slide of breast malignant tumor (stained with HE) seen in different magnification factors: (a) 40, (b) 100, (c) 200, and (d) 400.}
  \label{fig:fig1}
\end{figure}

\subsection{The visualization methods}
In this study, we utilized the Gradient \cite{Ref6} as well as LRP-Z and LRP-Epsilon \cite{Ref8} visualization methods. These methods provide saliency maps that indicate the influence of each pixel in the decision-making process of the CNN. 
In fact, the Gradient method constructs the heatmap by backpropagating the gradients of the output with respect to the input image through the layers of the CNN. This process allows us to highlight the regions that have the most impact on the final prediction. On the other hand, the LRP methods utilize designed propagation rules for each layer to propagate the prediction backward. These rules distribute the relevance values throughout the network, providing insights into the contribution of different layers and features to the overall decision.

Based on the heatmap generated by the visualization methods, we selected the pixels that have an influence on the decision of the CNN. For this selection process, we tested three different methods: Bins, K-means \cite{Ref18}, and MeanShift \cite{Ref19}. These selection methods allowed us to identify and isolate the regions of interest that significantly contribute to the CNN's decision-making process. By applying these methods, we aimed to gain a better understanding of the specific areas and patterns within the images that play a crucial role in the classification of malignancy.
 
\begin{itemize}
	\item Bins: the heatmaps are normalized in the interval [0; 1], then 10 bins were built for each heatmap.\\ 
	\item K-means: is a popular clustering algorithm with local optimization, the objective of K-means is to group similar data points together, we have used 10  clusters.\\
	\item	Mean shift: is a non parametric clustering technique which does not require prior knowledge of the number of clusters, and does not constrain the shape of the clusters. In this method we have selected the 10 first clusters.
\end{itemize}
 \section{Experimental results}
 \subsection{Quantitatively evaluation of visualization methods}
Firstly, we ordered the pixels based on their importance in the classification process using the selection methods. Secondly, we obscured the selected pixels in the input image by placing a black square over them. Finally, we applied the classification and observed the effect on accuracy. If the accuracy of classification decreases when the selected pixels are occluded, it indicates that those pixels are relevant to the classification process.

For these experiments, we employed the VGG19 architecture, and the performance of the classification was evaluated using the area under the receiver operating characteristic curve (AUC). This metric provides a comprehensive measure of the model's ability to discriminate between benign and malignant cases and allows for the comparison of different approaches and techniques used in the study.

In our approach, the best visualization method is the one that yields a low AUC value, as it indicates a higher degree of relevance in the classification process.
\\Using the Bins method, we obtained an AUC of 0.80 for the Gradient visualization method and 0.60 for both LRP-Z and LRP-Epsilon methods.
\\When applying the K-means method, the results were very close, with an AUC of 0.57 for the Gradient method and 0.53 for both LRP-Z and LRP-Epsilon methods.
\\Notably, the Mean Shift method outperformed the previous methods, with an AUC of 0.35 for the Gradient visualization and 0.51 for both LRP-Z and LRP-Epsilon methods.
\\Therefore, for the remainder of our experiments, we utilized the Gradient visualization method to produce the visualization map, and the Mean Shift method was employed to select salient pixels. These methods demonstrated promising performance in terms of identifying relevant features for the classification of breast histological images.

\begin{figure}[hbt!]
 \centering
  \includegraphics[scale=0.6]{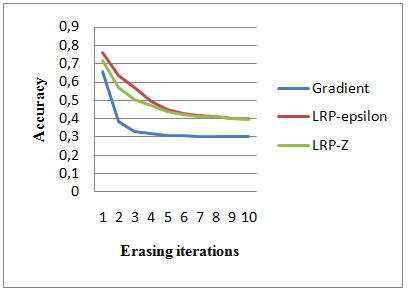}
  \caption{The erasing curve of the visualization algorithms for the 40X magnification.}
  \label{fig:fig2}
\end{figure}

Figure \ref{fig:fig2} illustrates the quantitative evaluation of the visualization methods at a magnification of 40X. The evaluation process involved erasing clusters of salient pixels based on their values in descending order. The curves of the Gradient method show that the accuracy decreases when the first relevant clusters are erased. However, it is interesting to note that the accuracy remains relatively high even in the first and second erasing clusters, which still contain salient pixels. This can be attributed to the small size of these erased clusters, which contain only a few pixels.
Furthermore, it is worth mentioning that the accuracy stabilizes from the fourth to the tenth erasing iteration. In other words, even when more pixels are erased, the accuracy remains consistent. This suggests that the most important regions are captured in these erasing iterations, and further erasing does not significantly impact the classification accuracy.
\subsection{Clinical evaluation of visualization method}
In this section, our focus is on highlighting the regions utilized by the CNN and the regions identified by the pathologist for the classification of histological images. We present a series of images at various magnification levels to facilitate our analysis.

Figures 3 to 5 showcase a sequence of images, numbered from 1 to 10, where the regions deemed important for classification have been occluded using black color. The occlusion is performed in accordance with the order of pixel importance in the classification process. Image number 11 displays the annotations provided by the expert pathologist indicating their respective levels of importance for classification. The regions encompassed by the red color represent the most significant and relevant areas according to the pathologist's classification. Following that, there are regions encompassed by orange, which are of secondary importance. Finally, the regions encompassed by blue hold a lower level of significance in the pathologist's classification.
\\The final image represents the original, unaltered image.
\\By examining these images, we can gain insights into the specific regions and features that contribute to the classification decisions made by both the CNN and the pathologist.
\\From all the cases that were tested, the initial three sequential images display only a small number of erased pixels, which are scattered throughout the image. The regions of interest are formed more and more in images 4-6.

In Figure 3, we examine a case of malignant ductal carcinoma, specifically at a magnification size of 40X. This type of cancer originates in the milk ducts of the breast. As the disease progresses, the cancer cells penetrate the walls of the ducts and invade the surrounding breast tissue.
\\The occluded regions in images 4-6 show similarities to the regions annotated by the pathologist as red and orange. Moreover, these regions become more prominent and expansive in images 7-10. As we discussed in Section 4, the occluded regions in the fourth, fifth, and sixth iterations represent the most crucial regions in the classification process. This alignment between the occluded regions and the pathologist's annotations further validates the importance and relevance of these regions for accurate classification.

In Figure 4, we examined an image of malignant ductal carcinoma at a magnification of 100X. The pathologist utilized all the available information in the image to classify it as malignant. However, it is evident that there are significant regions in the image that were not utilized by the CNN for classification. This disparity highlights the potential limitations of the CNN in capturing and utilizing all the relevant information present in the image for accurate classification.

In Figure 5, we examined a benign case of adenosis type at a magnification of 400X. The pathologist relied on the presence of epithelial and myoepithelial cells, as indicated by the red curve, to make the classification. However, the CNN utilized not only the epithelial and myoepithelial cells but also the connective tissue cells in its classification. This difference in the regions used by the pathologist and the CNN can explain the challenges faced by the CNN in distinguishing between epithelial, myoepithelial, and connective tissue cells, which are visually similar in the image.
Adenosis is a hyperplasia characterized by an increase in the number of glandular components, primarily affecting the lobular units. According to the pathologist's expertise, the identification of adenosis symptoms is more significant at 10X and 40X magnifications rather than at 400X magnification. This is because the image at 400X magnification reveals more intricate details of the nucleus, nucleolus, and chromatin, which can resemble carcinoma in situ. This discrepancy in the level of detail observed at different magnifications may explain the difficulty encountered by the CNN in accurately identifying the correct region of interest.

Despite the apparent simplicity of classifying histological images as benign or malignant, the differences observed in the identified regions emphasize that there are variations in the reasoning employed by pathologists and CNNs.
\newpage
\begin{figure}[!ht]
  \centering
  \begin{subfigure}[b]{0.3\linewidth}
    \includegraphics[width=\linewidth]{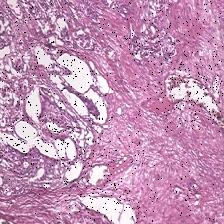}
    \caption{(1)}
  \end{subfigure}
  \begin{subfigure}[b]{0.3\linewidth}
    \includegraphics[width=\linewidth]{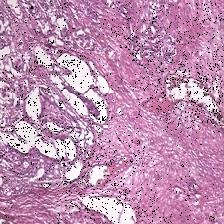}
    \caption{(2)}
  \end{subfigure}
  \begin{subfigure}[b]{0.3\linewidth}
    \includegraphics[width=\linewidth]{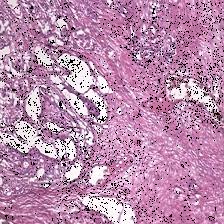}
    \caption{(3)}
  \end{subfigure}
  \begin{subfigure}[b]{0.3\linewidth}
    \includegraphics[width=\linewidth]{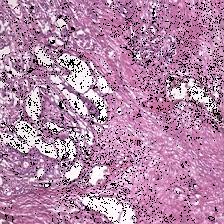}
    \caption{(4)}
  \end{subfigure}
  \begin{subfigure}[b]{0.3\linewidth}
    \includegraphics[width=\linewidth]{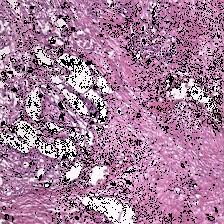}
    \caption{(5)}
  \end{subfigure}
  \begin{subfigure}[b]{0.3\linewidth}
    \includegraphics[width=\linewidth]{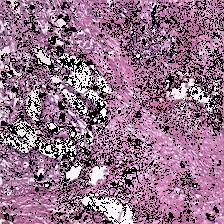}
    \caption{(6)}
  \end{subfigure}
  \begin{subfigure}[b]{0.3\linewidth}
    \includegraphics[width=\linewidth]{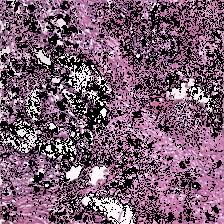}
    \caption{(7)}
  \end{subfigure}
  \begin{subfigure}[b]{0.3\linewidth}
    \includegraphics[width=\linewidth]{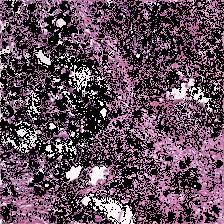}
    \caption{(8)}
  \end{subfigure}
  \begin{subfigure}[b]{0.3\linewidth}
    \includegraphics[width=\linewidth]{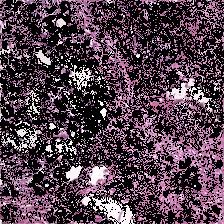}
    \caption{(9)}
  \end{subfigure}
  \begin{subfigure}[b]{0.29\linewidth}
    \includegraphics[width=\linewidth]{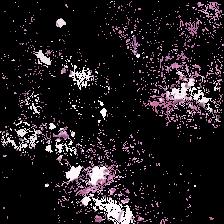}
    \caption{(10)}
  \end{subfigure}
    \begin{subfigure}[b]{0.31\linewidth}
    \includegraphics[width=\linewidth]{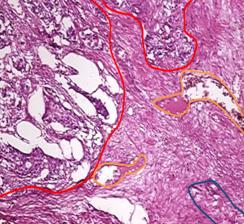}
    \caption{(11)}
  \end{subfigure}
  \begin{subfigure}[b]{0.31\linewidth}
    \includegraphics[width=\linewidth]{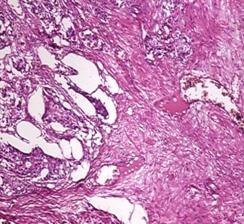}
    \caption{(12)}
  \end{subfigure}
    \caption{images (1-10) containing occluding regions, image (11) contains the pathologist annotation, image (12) represent the original image SOB-M-DC-14-16716-40-01011}
  \label{fig:fig3}
\end{figure}
\newpage
\begin{figure}[!ht]
  \centering
  \begin{subfigure}{0.3\linewidth}
    \includegraphics[width=\linewidth]{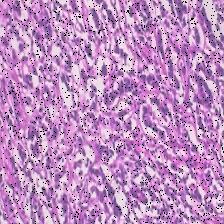}
    \caption{(1)}
  \end{subfigure}
  \begin{subfigure}{0.3\linewidth}
    \includegraphics[width=\linewidth]{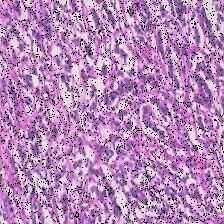}
    \caption{(2)}
  \end{subfigure}
  \begin{subfigure}{0.3\linewidth}
    \includegraphics[width=\linewidth]{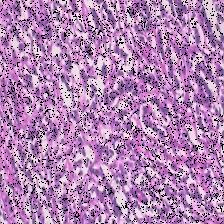}
    \caption{(3)}
  \end{subfigure}
  \begin{subfigure}{0.3\linewidth}
    \includegraphics[width=\linewidth]{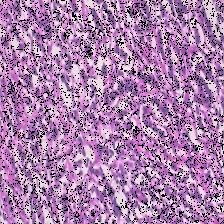}
    \caption{(4)}
  \end{subfigure}
  \begin{subfigure}{0.3\linewidth}
    \includegraphics[width=\linewidth]{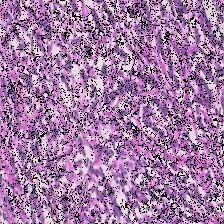}
    \caption{(5)}
  \end{subfigure}
  \begin{subfigure}{0.3\linewidth}
    \includegraphics[width=\linewidth]{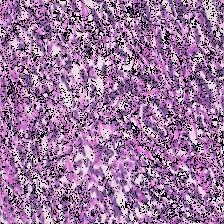}
    \caption{(6)}
  \end{subfigure}
  \begin{subfigure}{0.3\linewidth}
    \includegraphics[width=\linewidth]{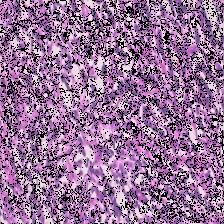}
    \caption{(7)}
  \end{subfigure}
  \begin{subfigure}{0.3\linewidth}
    \includegraphics[width=\linewidth]{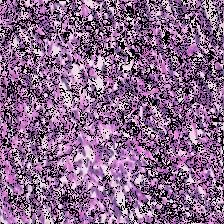}
    \caption{(8)}
  \end{subfigure}
  \begin{subfigure}{0.3\linewidth}
    \includegraphics[width=\linewidth]{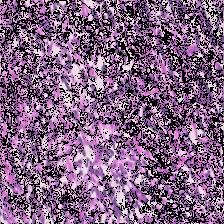}
    \caption{(9)}
  \end{subfigure}
  \begin{subfigure}{0.29\linewidth}
    \includegraphics[width=\linewidth]{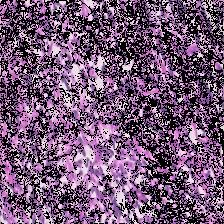}
    \caption{(10)}
  \end{subfigure}
    \begin{subfigure}{0.31\linewidth}
    \includegraphics[width=\linewidth]{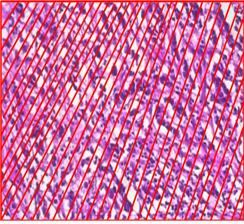}
    \caption{(11)}
  \end{subfigure}
  \begin{subfigure}{0.31\linewidth}
    \includegraphics[width=\linewidth]{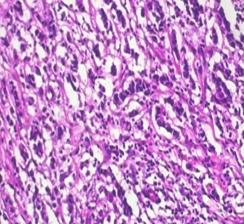}
    \caption{(12)}
  \end{subfigure}
      \caption{images (1-10) containing occluding regions, image (11) contains the pathologist annotation, image (12) represent the original image SOB-M-DC-14-11520-100-019}
  \label{fig:fig4}
\end{figure}
\begin{figure}[!ht]
  \centering
  \begin{subfigure}[b]{0.3\linewidth}
    \includegraphics[width=\linewidth]{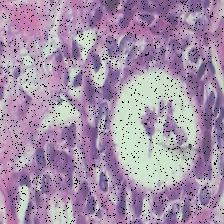}
    \caption{(1)}
  \end{subfigure}
  \begin{subfigure}[b]{0.3\linewidth}
    \includegraphics[width=\linewidth]{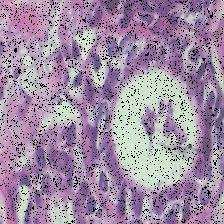}
    \caption{(2)}
  \end{subfigure}
  \begin{subfigure}[b]{0.3\linewidth}
    \includegraphics[width=\linewidth]{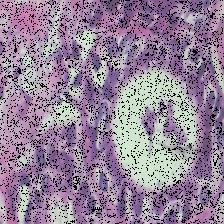}
    \caption{(3)}
  \end{subfigure}
  \begin{subfigure}[b]{0.3\linewidth}
    \includegraphics[width=\linewidth]{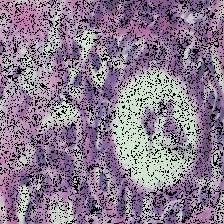}
    \caption{(4)}
  \end{subfigure}
  \begin{subfigure}[b]{0.3\linewidth}
    \includegraphics[width=\linewidth]{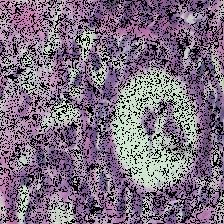}
    \caption{(5)}
  \end{subfigure}
  \begin{subfigure}[b]{0.3\linewidth}
    \includegraphics[width=\linewidth]{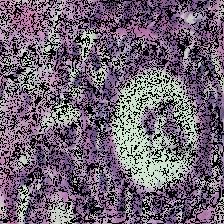}
    \caption{(6)}
  \end{subfigure}
  \begin{subfigure}[b]{0.3\linewidth}
    \includegraphics[width=\linewidth]{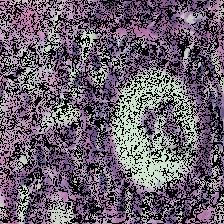}
    \caption{(7)}
  \end{subfigure}
  \begin{subfigure}[b]{0.3\linewidth}
    \includegraphics[width=\linewidth]{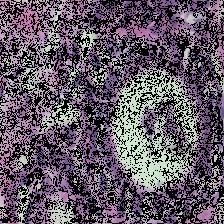}
    \caption{(8)}
  \end{subfigure}
  \begin{subfigure}[b]{0.3\linewidth}
    \includegraphics[width=\linewidth]{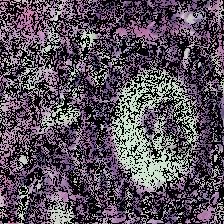}
    \caption{(9)}
  \end{subfigure}
  \begin{subfigure}[b]{0.29\linewidth}
    \includegraphics[width=\linewidth]{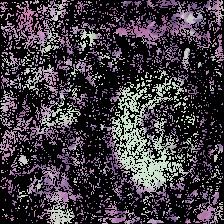}
    \caption{(10)}
  \end{subfigure}
    \begin{subfigure}[b]{0.31\linewidth}
    \includegraphics[width=\linewidth]{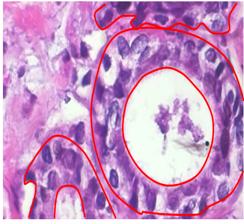}
    \caption{(11)}
  \end{subfigure}
  \begin{subfigure}[b]{0.31\linewidth}
    \includegraphics[width=\linewidth]{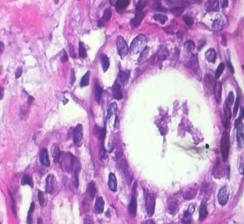}
    \caption{(12)}
  \end{subfigure}
   \caption{ images (1-10) containing occluding regions, image (11) contains the pathologist annotation, image (12) represent the original image SOB-B-A-14-22549G-400-009}
  \label{fig:fig6}
\end{figure}

\section {Conclusion}

In this study, our main focus was on visualizing the regions of interest utilized by the CNN for classifying histological images as either benign or malignant. We initially experimented with two visualization methods, namely Gradient and LRP. To identify the important pixels, we tested three selection methods: Bins, K-means, and Mean shift. Based on the obtained results, we chose Gradient as our visualization method and Mean shift for selecting salient pixels.

The comparison between the regions identified by the CNN and the pathologist revealed that there were some regions used by the CNN for classification that were not considered significant by the pathologist, and vice versa. However, we did find certain regions that were commonly identified by both approaches.

By applying the visualization methods, we were able to gain insights into the behavior of the CNN and better understand its classification decisions. As part of our future work, we intend to fine-tune the parameters of the CNN to align the regions of interest used by the CNN with the annotations provided by the pathologist.

\vspace{12pt}
\end{document}